# Modifying the Symbolic Aggregate Approximation Method to Capture Segment Trend Information

Muhammad Marwan Muhammad Fuad

Coventry University
Coventry CV1 5FB, UK
ad0263@coventry.ac.uk

**Abstract:** The Symbolic Aggregate approXimation (SAX) is a very popular symbolic dimensionality reduction technique of time series data, as it has several advantages over other dimensionality reduction techniques. One of its major advantages is its efficiency, as it uses precomputed distances. The other main advantage is that in SAX the distance measure defined on the reduced space lower bounds the distance measure defined on the original space. This enables SAX to return exact results in query-by-content tasks. Yet SAX has an inherent drawback, which is its inability to capture segment trend information. Several researchers have attempted to enhance SAX by proposing modifications to include trend information. However, this comes at the expense of giving up on one or more of the advantages of SAX. In this paper we investigate three modifications of SAX to add trend capturing ability to it. These modifications retain the same features of SAX in terms of simplicity, efficiency, as well as the exact results it returns. They are simple procedures based on a different segmentation of the time series than that used in classic-SAX. We test the performance of these three modifications on 45 time series datasets of different sizes, dimensions, and nature, on a classification task and we compare it to that of classic-SAX. The results we obtained show that one of these modifications manages to outperform classic-SAX and that another one slightly gives better results than classic-SAX.

**Keywords:** Classification, SAX, Time series mining.

## 1 Introduction

Several medical, financial, and weather forecast activities produce data in the form of measurements recorded over a period of time. This type of data is known as time series. Time series data mining has witnessed substantial progress in the last two decades because of the variety of applications to this data type. It is estimated that much of today's data come in the form of time series [17].

There are a number of common time series data mining tasks, such as classification, clustering, query-by-content, anomaly detection, motif discovery, prediction, and others [7]. The key to performing these tasks effectively and efficiently is to have a high-quality representation of these data to capture their main characteristics.

Several time series representation methods have been proposed. The most common ones are *Discrete Fourier Transform* (DFT) [1] and [2], *Discrete Wavelet Transform* (DWT) [5], *Singular Value Decomposition* (SVD) [12], *Adaptive Piecewise Constant Approximation* (APCA) [11], *Piecewise Aggregate Approximation* (PAA) [10] and [23], *Piecewise Linear Approximation* (PLA) [18], and *Chebyshev Polynomials* (CP) [4].

Another very popular time series representation method, which is directly related to this paper, is the *Symbolic Aggregate approXimation* method (SAX) [13], [14]. The reason behind its popularity is its simplicity and efficiency, as it uses precomputed lookup tables. Another reason is its ability to return exact results in query-by-content tasks. The drawback of SAX is that during segmentation and symbolic representation, the trend information of the segments is lost, which results in lower-quality representation and less pruning power.

Several papers have spotted this drawback in SAX and there have been a few attempts to remedy it. All of them, however, had to sacrifice one, or even both, of the main advantages of SAX; its simplicity and its ability to return exact results in query-by-content tasks.

In this paper we propose three modifications of SAX that attempt to capture, to a certain degree, the trend information of segments. The particularity of our modifications is that they retain the two main advantages of the original SAX, which we call *classic-SAX* hereafter, as our modifications have the same simplicity and require exactly the same computational cost as classic-SAX. They also return exact results in query-by-content tasks.

We conduct classification experiments on a wide variety of time series datasets obtained from the time series archive to validate our method. The results were satisfying. It is important to mention here that we are not expecting the results to "drastically" outperform those of classic-SAX given that we kept exactly the same simplicity and efficiency of classic-SAX, which was the objective of our method while we were developing it.

The rest of this paper is organized as follows; Section 2 is a background section. The new method with its three versions is presented in Section 3, and is validated experimentally in Section 4. We conclude with Section 5.

## 2    Background

Time series data mining has witnessed increasing interest in the last two decades. The size of time series databases has also grown considerably. Because of the high-dimensionality and high feature correlation of time series data, representation methods, which are dimensionality reduction techniques, have been proposed as a means to perform data mining tasks on time series data.

The GEMINI framework [8] reduces the dimensionality of the time series from a point in an $n$-dimensional space into a point in an $m$-dimensional space (some methods use several low-dimensional spaces like [19]), where $m \ll n$. The similarity measure defined on the $m$-dimensional space is said to be *lower bounding* of the original similarity measure defined on the $n$-dimensional space if:

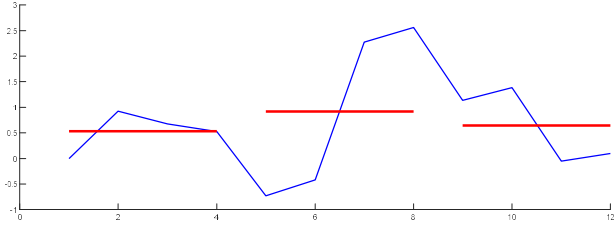

**Fig. 1.** PAA representation

$$d^m(\acute{S}, \acute{T}) \leq d^n(S, T) \tag{1}$$

where $\acute{S}$ and $\acute{T}$ are the representations of time series $S$ , $T$, respectively, on the $m$-dimensional space. Applying the GEMINI framework guarantees that the similarity search queries will not produce false dismissals.

One of the most popular dimensionality reduction techniques of time series is the *Piecewise Aggregate Approximation* (PAA) [10], [23]. PAA divides a time series $S$ of $n$-dimensions into $m$ equal-sized segments (words) and maps each segment to a point in a lower $m$-dimensional space, where each point in the reduced space is the mean of values of the data points falling within this segment. The similarity measure given in the following equation:

$$d^m(S, T) = \sqrt{\frac{n}{m}} \sqrt{\sum_{i=1}^{m} (\bar{s}_i - \bar{t}_i)^2} \tag{2}$$

is defined on the $m$-dimensional space, where $\bar{s}_i$ , $\bar{t}_i$ are the averages of the points in segment $i$ in $S$ , $T$, respectively.

PAA is the basis for another popular and very efficient time series dimensionality reduction technique, which is the *Symbolic Aggregate approXimation* – SAX [13], [14]. SAX is based on the assumption that normalized time series have a Gaussian distribution, so by determining the locations of the breakpoints that correspond to a particular alphabet size, chosen by the user, one can obtain equal-sized areas under the Gaussian curve. SAX is applied to normalized time series in three steps as follows:

1- The dimensionality of the time series is reduced using PAA.

2- The resulting PAA representation is discretized by determining the number and locations of the breakpoints. The number of the breakpoints $nrBreakPoints$ is related to the alphabet size $alphabetSize$; i.e. $nrBreakPoints = alphabetSize - 1$. As for their locations, they are determined, as mentioned above, by using Gaussian lookup tables. The interval between two successive breakpoints is assigned to a symbol of the alphabet, and each segment of PAA that lies within that interval is discretized by that symbol.

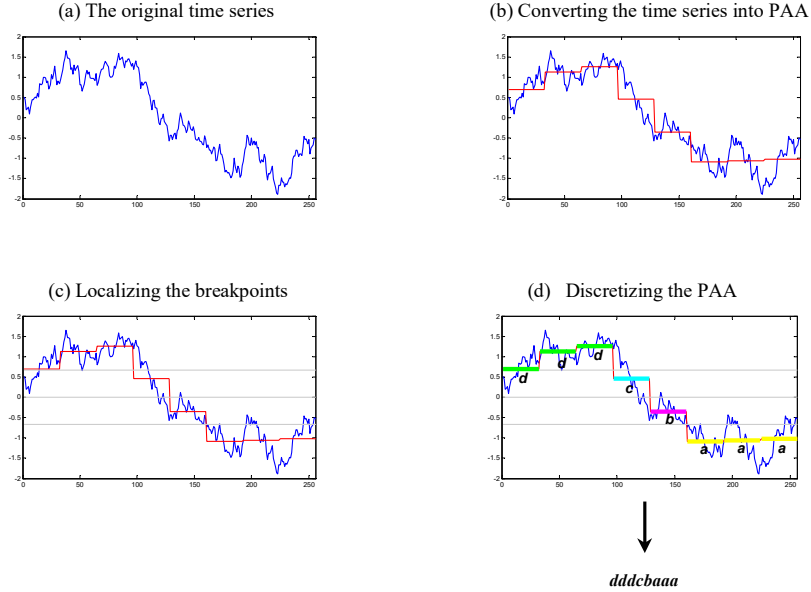

**Fig. 2.** The different steps of SAX

3- The last step of SAX is using the following similarity measure:

$$MINDIST(\hat{S}, \hat{T}) = \sqrt{\frac{n}{m}} \sqrt{\sum_{i=1}^{m} \left( dist(\hat{s}_i, \hat{t}_i) \right)^2} \tag{3}$$

Where $n$ is the length of the original time series, $m$ is the number of segments, $\hat{S}$ and $\hat{T}$ are the symbolic representations of the two time series $S$ and $T$, respectively, and where the function $dist(\ )$ is implemented by using the appropriate lookup table. For instance, the lookup table for an alphabet size of 3 is the one shown in Table 1. Fig. 2 illustrates the different steps of SAX.

It is proven in [13], [14] that the similarity distance defined in equation (3) is a lower bound of the Euclidean distance applied in the original space of time series.

Despite its popularity, SAX has a main drawback that it inherits from PAA, which is its inability to capture trend information during discretization. For example, the two segments:

**Table 1.** The lookup table of *MINDIST* for alphabet size = 3.

|   | a | b | c |
|---|---|---|---|
| a | 0 | 0 | 0.86 |
| b | 0 | 0 | 0 |
| c | 0.86 | 0 | 0 |

$S_1 = [-6, -1, +7, +8]$ and $S_2 = [+9, +3, +1, -5]$ have the same PAA coefficient which is $+2$ so their SAX representation is the same, although, as we can clearly see, their trends are completely different.

Several researchers have reported this flaw in SAX and attempted to propose different solutions to handle it. In [16] the authors present 1d-SAX which incorporates trend information by applying linear regression to each segment of the time series and then discretizing it using a symbol that incorporates both trend and mean of the segment at the same time. The method applies a complex representation scheme. It also includes additional parameters that need training. In [24] another extension of SAX that takes into account both trend and value is proposed. This method first transforms each real value into a binary bit. This representation is used in a first pruning step. In the second step the trend is represented as a binary bit. This method requires two passes through the data which makes it inefficient for large datasets. [9] proposes to consider both value and trend information by combining PAA with an extended version of the clipping technique [21]. However, obtaining good results requires prior knowledge of the data.

In addition to all these drawbacks we mentioned above of each of these methods, which attempt to enable SAX to capture trend information, they all have two main disadvantages: a) whereas the main merit of SAX is its simplicity, all these methods are more/ much more complex than classic-SAX. They require training of parameters and/or preprocessing steps. b) the main disadvantage of these methods is that the distance measure they propose on the low-dimension space does not lower bound the distance measure applied in the original space of the raw data. As a consequence, they do not return all the answers of a query-by-content task, unlike classic-SAX which returns exact results because it applies a lower bounding distance.

## 3  The Proposed Method

Our method aims to integrate trend information into classic-SAX all by fulfilling the following requirements:

i. Conserving the simplicity of classic-SAX.
ii. Not adding any additional steps.
iii. Not adding any additional parameters.
iv. Keeping the lower bounding condition of classic-SAX.

Our method is based on the two following remarks:

a. Trend can be better captured by including points that are farther apart than if they were adjacent.
b. Very interestingly, equation (2) on which PAA is based, and consequently equation (3), does not stipulate that the points of a segment should be adjacent. It only stipulates that the points should be calculated once only in the summation, and of course, that each point of one time series should be compared with its counter point in the other time series. In simple words, if, say we are using a compression ratio of 4:1, i.e. each segment of four points in the original space is represented by the mean of these four points, then these four points do not have to be adjacent for equation (2) to be valid.

Based on these two remarks, we propose the following three methods to segment the time series and calculate the means in PAA, and consequently in SAX:

**1-Overlap-SAX:** In this version, the end points of each segment are swapped with the end points of the two neighboring segments; i.e. the first point of segment $m$ is swapped with the end point of segment $m - 1$, and the end point of segment $m$ is swapped with the first point of segment $m + 1$. As for the first and last segments of the time series, only the last, first, respectively, points are swapped.

For example, for a time series of 16 points and for a compression ratio of 4:1, according to overlap-SAX this time series is divided into four segments as follows:
$\langle x_1, x_2, x_3, x_5 \rangle, \langle x_4, x_6, x_7, x_9 \rangle, \langle x_8, x_{10}, x_{11}, x_{13} \rangle, \langle x_{12}, x_{14}, x_{15}, x_{16} \rangle$

The next steps are identical to those of classic-SAX, i.e. the mean of each segment is calculated then discretized using the corresponding lookup table, and finally equation (3) is applied.

**2-Intertwine-SAX:** In this version, as the name suggests, each other point belongs to one segment, the next segment consists of the points we skipped when constructing the previous segment, and so on. For the same example as above, in intertwine-SAX this 16-point time series is divided into four segments as follows:
$\langle x_1, x_3, x_5, x_7 \rangle, \langle x_2, x_4, x_6, x_8 \rangle, \langle x_9, x_{11}, x_{13}, x_{15} \rangle, \langle x_{10}, x_{12}, x_{14}, x_{16} \rangle$

The rest of the steps are identical to those of classic-SAX.

**3-Split-SAX:** In this version, two successive points are assigned to a segment $m$, then two successive points are skipped, then the two successive points are assigned to $m$ too, segment $m + 1$ consists of the skipped points when constructing $m$ in addition to the two successive points following segment $m$, and so on. The 16-point time series in our example is divided into four segments in spilt-SAX as follows:
$\langle x_1, x_2, x_5, x_6 \rangle, \langle x_3, x_4, x_7, x_8 \rangle, \langle x_9, x_{10}, x_{13}, x_{14} \rangle, \langle x_{11}, x_{12}, x_{15}, x_{16} \rangle$.
This version attempts to capture trend on a wider range than the two previous ones.

As we can see, the three versions do not add any additional complexity to classic-SAX, they conserve all its main characteristics, mainly its simplicity. They also lower bound the original distance as we showed in remark (b) above. In fact, even coding each of these versions requires only a very simple modification of the original code of classic-SAX.

## 4  Experiments

Classification is one of the main tasks in data mining. In classification we have categorical variables which represent classes. The task is to assign class labels to the dataset according to a model learned during a learning stage on a training dataset, where the class labels are known. When given new data, the algorithm aims to classify these data based on the model acquired during the training stage, and later applied to a testing dataset.

There are a number of classification models, the most popular of which is *k-nearest-neighbor* (*k*NN). In this model the object is classified based on the *k* closest objects in its neighborhood. A special case of particular interest is when $k = 1$, which we use in this paper.

Time series classification has several real world applications such as health care [15], security [22], food safety [20], and many others.

The performance of classification algorithms can be evaluated using different methods. One of the widely used ones is *leave-one-out cross-validation* (LOOCV) (also known as *N-fold cross-validation*, or *jack-knifing*), where the dataset is divided into as many parts as there are instances, each instance effectively forming a test set of one. N classifiers are generated, each from $N - 1$ instances, and each is used to classify a single test instance. The classification error is then the total number of misclassified instances divided by the total number of instances [3].

We compared the performance of the three versions we presented in Section 3; overlap-SAX, intertwine-SAX, and split-SAX, to that of classic-SAX in a 1NN classification task on 45 time series datasets of different sizes and dimensions available at the *UCR Time Series Classification Archive* [6]. Each dataset in this archive is divided into a training dataset and a testing dataset. The dimension (length) of the time series on which we conducted our experiments varied between 24 (ItalyPowerDemand) and 1882 (InlineSkate). The size of the training sets varied between 16 instances (DiatomSizeReduction) and 560 instances (FaceAll). The size of the testing sets varied between 20 instances (BirdChicken), (BeetleFly) and 3840 instances (ChlorineConcentration). The number of classes varied between 2 (Gun-Point), (ECG200), (Coffee), (ECGFiveDays), (ItalyPowerDemand), (MoteStrain), (TwoLeadECG), (BeetleFly), (BirdChicken), (Strawberry), (Wine), and 37 (Adiac).

Applying the different versions of SAX comprises two stages. In the training stage we obtain the alphabet size that yields the minimum classification error for that version of SAX on the training dataset. Then in the testing stage we apply the investigated version of SAX (classic, overlap, intertwine, split) to the corresponding testing dataset, using the alphabet size obtained in the training stage, to obtain the classification error on the testing dataset.

In Table 2 we show the results of the experiments we conducted. The best result (the minimum classification error) for each dataset is shown in underlined, boldface printing. In the case where all versions give the same classification error for a certain dataset, the result is shown is italics.

There are several measures used to evaluate the performance of time series classification methods. In this paper we choose a simple and widely used one, which is to count how many datasets on which the method gave the best performance. Of the four versions tested, the performance of overlap-SAX is the best as it yielded the minimum classification error on 15 datasets. We have to say these were the results we were expecting. In fact, while we were developing the methods presented in this paper overlap-SAX was the first version we thought of as it mainly focusses on the principle information of each segment and only adds the "extra" required information to capture the trend by swapping the end points of each segment with the neighboring segments.

**Table 2.** The 1NN classification error of classic-SAX, overlap-SAX, intertwine-SAX, and split-SAX The best result for each dataset is shown in underlined, boldface printing. Results shown in italics are those where the four versions gave the same classification error.

| Dataset | Method | | | |
|---|---|---|---|---|
| | classic-SAX | overlap-SAX | intertwine-SAX | split-SAX |
| synthetic_control | **_0.023333_** | 0.05 | 0.07 | 0.063333 |
| Gun_Point | 0.14667 | 0.14 | 0.19333 | **_0.13333_** |
| CBF | 0.075556 | 0.095556 | **_0.06_** | 0.085556 |
| FaceAll | 0.30473 | **_0.27811_** | 0.3 | 0.31953 |
| OSULeaf | 0.47521 | 0.4876 | 0.48347 | **_0.47107_** |
| SwedishLeaf | **_0.2528_** | 0.2656 | 0.2768 | 0.272 |
| Trace | 0.37 | **_0.28_** | 0.32 | 0.29 |
| FaceFour | 0.22727 | 0.20455 | **_0.17045_** | 0.20455 |
| Lighting2 | **_0.19672_** | 0.21311 | 0.37705 | 0.40984 |
| Lighting7 | 0.42466 | 0.39726 | 0.43836 | **_0.38356_** |
| ECG200 | 0.12 | **_0.11_** | 0.12 | 0.13 |
| Adiac | 0.86701 | 0.86701 | 0.86445 | **_0.85678_** |
| yoga | **_0.18033_** | 0.18233 | 0.186 | 0.18433 |
| FISH | 0.26286 | 0.22286 | **_0.21714_** | 0.24571 |
| Plane | _0.028571_ | _0.028571_ | _0.028571_ | _0.028571_ |
| Car | **_0.26667_** | 0.28333 | 0.3 | 0.3 |
| Beef | 0.43333 | **_0.4_** | 0.43333 | 0.43333 |
| Coffee | 0.28571 | **_0.25_** | 0.28571 | **_0.25_** |
| OliveOil | _0.83333_ | _0.83333_ | _0.83333_ | _0.83333_ |
| CinC_ECG_torso | **_0.073188_** | **_0.073188_** | 0.075362 | 0.12246 |
| ChlorineConcentration | 0.58203 | **_0.54661_** | 0.57891 | 0.5625 |
| DiatomSizeReduction | 0.081699 | 0.088235 | 0.084967 | **_0.078431_** |
| ECGFiveDays | 0.14983 | **_0.13821_** | 0.18815 | 0.17886 |
| FacesUCR | 0.24244 | **_0.19366_** | 0.23024 | 0.24537 |
| Haptics | 0.64286 | 0.63636 | **_0.63312_** | 0.64935 |
| InlineSkate | 0.68 | 0.67818 | 0.67455 | **_0.67273_** |
| ItalyPowerDemand | 0.19242 | **_0.16035_** | 0.31487 | 0.25462 |
| MALLAT | 0.14328 | 0.14456 | **_0.14286_** | 0.15309 |
| MoteStrain | 0.21166 | 0.19249 | 0.1885 | **_0.17812_** |
| SonyAIBORobotSurface | 0.29784 | **_0.28785_** | 0.32612 | 0.2995 |
| SonyAIBORobotSurfaceII | **_0.14376_** | 0.19973 | 0.32529 | 0.15845 |
| Symbols | **_0.10251_** | 0.10452 | 0.10553 | 0.10553 |
| TwoLeadECG | 0.30904 | 0.34241 | 0.30729 | **_0.29939_** |
| InsectWingbeatSound | 0.44697 | 0.45253 | 0.44949 | **_0.44141_** |
| ArrowHead | 0.24571 | 0.22857 | 0.22857 | **_0.22286_** |
| BeetleFly | _0.25_ | _0.25_ | _0.25_ | _0.25_ |
| BirdChicken | **_0.35_** | 0.4 | 0.4 | 0.4 |
| Ham | **_0.34286_** | 0.39048 | 0.41905 | 0.41905 |
| Herring | _0.40625_ | _0.40625_ | _0.40625_ | _0.40625_ |
| ToeSegmentation1 | 0.36404 | **_0.35088_** | 0.35965 | 0.39035 |
| ToeSegmentation2 | 0.14615 | 0.20769 | **_0.13846_** | 0.18462 |
| DistalPhalanxOutlineAgeGroup | 0.2675 | **_0.235_** | 0.5325 | 0.4425 |
| DistalPhalanxOutlineCorrect | 0.34 | 0.38333 | **_0.28333_** | 0.29 |
| DistalPhalanxTW | 0.2925 | **_0.2725_** | 0.3175 | 0.3275 |
| WordsSynonyms | 0.37147 | **_0.3652_** | 0.37147 | 0.37147 |
| | 10 | 15 | 7 | 11 |

The second best version is split-SAX, which gave the minimum classification error on 11 datasets. The third best version is classic-SAX, whose performance is very close to that of split-SAX, as it gave the minimum classification error in 10 of the tested datasets. The last version is intertwine-SAX which gave the minimum classification error in only 7 of the tested datasets. The four versions gave the same classification error on three datasets (Plane), (OliveOil), and (BeetleFly).

It is interesting to notice that even the least performing version gave better results than the others on certain datasets. This can be a possible direction of further research for the work we are presenting in this paper - to investigate why a certain version works better with a certain dataset.

We have to add that although the versions we presented in Section 3 apply a compression ratio of 4:1, which is the compression ratio used in classic-SAX, the extension to a different compression ratio is straightforward, except for split-SAX which requires a simple modification in the case where the segment length is an odd number.

## 5 Conclusion and Future Work

In this paper we presented three modifications of classic-SAX, a powerful time series dimensionality reduction technique. These modifications were proposed to enhance the performance of classic-SAX by enabling it to capture trend information. These modifications; overlap-SAX, intertwine-SAX, and split-SAX, were designed so that they retain the exact same features of classic-SAX as they all have the same efficiency, simplicity, and they use a lower bounding distance. We compared their performance with that of classic-SAX, and we showed how one of them, overlap-SAX, gives better results than classic-SAX, and another one, split-SAX, gives slightly better results than classic-SAX. This improvement, although not substantial, is still interesting as it did not require any additional pre-processing / post-processing steps, or any supplementary storage requirement. It only required a very simply modification that is based solely on segmenting the time series differently compared to classic-SAX.

There are two directions of future work, the first is to relate the trend of each segment to the global trend of the time series, so the trend of each segment should actually only capture how it "deviates" from the global trend.

The other direction is to explore new segmentation methods, whether when applied to SAX or to other time series dimensionality reduction techniques. We believe another indirect outcome of this paper is that it opens the door to considering new schemes for segmenting time series that do not necessarily focus on grouping adjacent points.